\documentclass[10pt,letterpaper]{article}

\usepackage{cogsci}
\usepackage{booktabs}

\cogscifinalcopy 

\usepackage{pslatex}
\usepackage{apacite}
\usepackage{natbib}

\usepackage[dvipsnames]{xcolor}
\usepackage{graphicx}
\usepackage{soul}

\usepackage{float}

\usepackage{hyperref}
\definecolor{coco1}{HTML}{D9E4EC}
\definecolor{coco2}{HTML}{B7CFDC}
\definecolor{coco3}{HTML}{6AABD2}
\definecolor{coco4}{HTML}{385E72}
\hypersetup{
    colorlinks=true,
    linkcolor=coco3,
    filecolor=coco4,      
    urlcolor=coco3,
    citecolor=coco3,
}

\newif\ifsubmit
\submitfalse
\ifsubmit
\newcommand{\ben}[1]{}
\newcommand{\ced}[1]{}
\newcommand{\todo}[1]{}
\newcommand{\tocheck}[1]{}
\newcommand{\ndg}[1]{}
\newcommand{\kg}[1]{}
\else
\fi

\title{Scaling up the think-aloud method}

\author{{\large \bf Daniel Wurgaft$^*$ } \\
  Psychology \\ Stanford University \\ \texttt{wurgaft@stanford.edu}
  \And {\large \bf Ben Prystawski$^*$} \\
  Psychology \\ Stanford University \\ \texttt{benpry@stanford.edu}
  \And {\large \bf Kanishk Gandhi} \\
  Computer Science \\ Stanford University
  \AND {\large \bf Cedegao E. Zhang} \\
  Brain and Cognitive Sciences \\ MIT
  \And {\large \bf Joshua B. Tenenbaum} \\
  Brain and Cognitive Sciences \\ MIT
  \And {\large \bf Noah D. Goodman} \\
  Psychology and Computer Science \\
  Stanford University
  }

\begin{document}

\maketitle

\begin{abstract}

The think-aloud method, where participants voice their thoughts as they solve a task, is a valuable source of rich data about human reasoning processes. Yet, it has declined in popularity in contemporary cognitive science, largely because labor-intensive transcription and annotation preclude large sample sizes. Here, we develop methods to automate the transcription and annotation of verbal reports of reasoning using natural language processing tools, allowing for large-scale analysis of think-aloud data. In our study, 640 participants thought aloud while playing the Game of 24, a mathematical reasoning task. We automatically transcribed the recordings and coded the transcripts as search graphs, finding moderate inter-rater reliability with humans. We analyze these graphs and characterize consistency and variation in human reasoning traces. Our work demonstrates the value of think-aloud data at scale and serves as a proof of concept for the automated analysis of verbal reports.

\textbf{Keywords:} 
think-aloud; verbal protocol; reasoning; problem solving; language models
\end{abstract}

\section{Introduction}

\renewcommand{\thefootnote}{*}
\footnotetext[0]{These authors contributed equally.}
\renewcommand{\thefootnote}{\arabic{footnote}}

Understanding how people reason is central to cognitive science, yet reasoning is notoriously elusive to study. Reasoning happens internally and unfolds over time, so studying it requires making inferences about the time course of a process we cannot observe directly. Throughout the field's history, cognitive scientists have developed methods to carefully infer thought processes from the data we \textit{can} observe directly. One such method, which was prominent in the formative years of the field, is the \textit{concurrent verbal protocol}, or \textit{think-aloud}, method \citep{newell_simon1972human,ericsson1993protocol,van1994think}.

In a think-aloud study, experimenters present a participant with a problem and instruct them to say whatever comes to mind as they solve it. The experimenters then study the transcript of what the participant said to infer the process they went through on their way to the solution \citep{ericsson1993protocol}. Classic studies applied this method to investigate reasoning in varied settings such as chess \citep{newell_simon1972human}, physics \citep{larkin1980expert}, and insight problems \citep{kaplan1990search}. Think-aloud data informed the development of early computational models of cognition, including the General Problem Solver \citep{newell1961gps}. The method provides rich data that elucidate the \textit{process} of reasoning, in addition to its outcomes. Therefore, think-aloud studies let us analyze reasoning at the algorithmic level, in the framework of \citet{marr1982vision}.

Despite its continued use in fields such as education and human-computer interaction \citep{van2017examining, hertzum2024concurrent}, the think-aloud method has largely fallen out of favor in contemporary cognitive science. Traditional think-aloud studies relied on sample sizes of only one or few participants, since coding transcripts requires substantial training and time \citep{newell_simon1972human,larkin1980expert}. As the field embraced large-scale experiments to enhance the robustness of findings and explore individual differences \citep[e.g.,][]{hartshorne2018critical, peterson2021using}, extensive manual coding of data became impractical.

The alternative to manual coding is an automated coding pipeline, which would allow the melding of process-level insights from think-aloud data with the large-scale experiments of contemporary cognitive science. 
Automatic coding of think-aloud data was originally proposed by \citet{waterman1971protocol}. However, the shortcomings of available artificial intelligence at the time limited the applicability of their system. Given recent developments in natural language processing (NLP)---particularly speech-to-text models \citep{radford2023robust} and large language models (LLMs) \citep{brown2020language}---new tools are available to study verbal reports. 

Therefore, we believe the time is right to revitalize the think-aloud method in cognitive science. Automated coding will enable researchers to process verbal reports from many participants, yielding data with both granularity and scale. Scale can also help mitigate concerns about the validity of think-aloud studies—specifically, that prompting participants to verbalize their thoughts may alter how they solve tasks \citep{wilson1994proper}. While some studies support this claim \citep{russo1989validity}, a more recent meta-analysis of 94 studies found no evidence that prompting to think aloud changes behavior \citep{fox2011procedures}. Still, large comparisons between reasoning behavior with and without verbalization are useful for understanding whether and how prompting to think aloud changes the distribution of final behavior. 

\begin{figure*}
    \centering
    \includegraphics[width=0.94\linewidth]{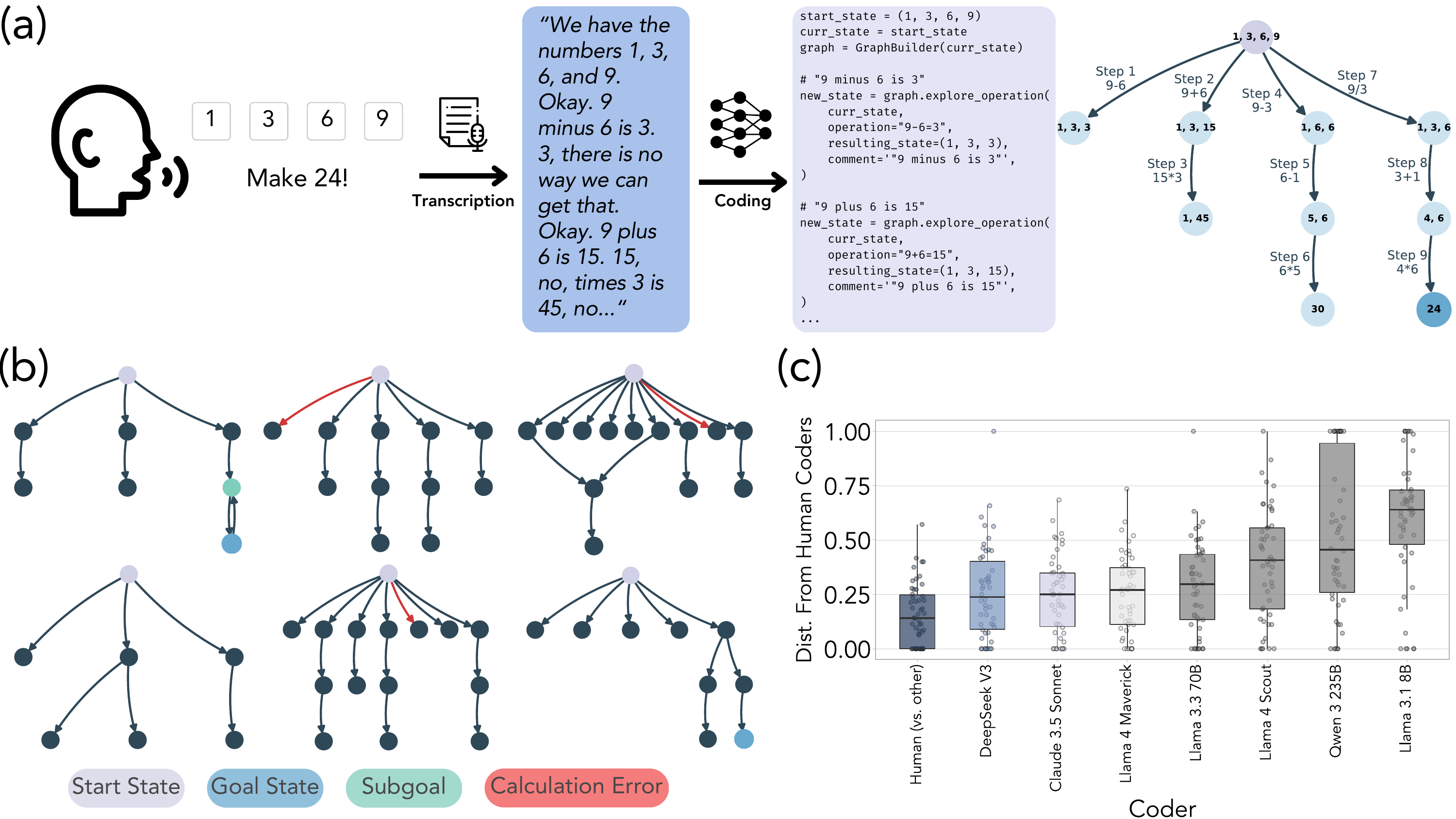}
    \vspace{-0.2cm}
    \caption{\textbf{Overview of the data processing pipeline.} (a) Participants solve the task while speaking aloud. We transcribe their data using a speech-to-text model and use an LLM to translate the transcripts into Python code that produces a graph representation of their search. (b) Example graphs for one problem in the dataset. (c)  Normalized graph edit distances between the graphs produced by human and language model coders. Lower edit distances indicate more agreement.}
    \label{fig:overview}
    
\end{figure*}

Several recent studies have used NLP tools to analyze verbal reports in risky choice and memory tasks \citep{ostrovsky2024verbal,mechera2024using,xie2023text2decision, xie2024evaluating}. These papers have provided valuable new ways to extract psychological information from unstructured audio and text. However, they focus on classifying whole utterances in simple settings \citep{ostrovsky2024verbal, xie2023text2decision, mechera2024using}. In contrast, many seminal think-aloud studies focused on producing \textit{process-level} insights in reasoning tasks that allow us to understand the individual steps taken when reasoning, in the spirit of \citet{newell1961gps}'s early cognitive modeling work.

In this work, we develop and validate an automated coding pipeline for processing think-aloud data in the context of a mathematical problem-solving task. We use the Game of 24, a simple yet rich arithmetic task that is well-suited to the study of reasoning. This game, along with its generalized version (also known as \href{https://en.wikipedia.org/wiki/Countdown_(game_show)}{Countdown}), has been used to study the development of mathematical reasoning \citep{tong2014card, van2017cognitive} and is a popular benchmark for evaluating LLMs' ability to reason \citep{yao2023tree,besta2024graph,gandhi2024stream}. We analyze nearly 5,000 verbal traces of reasoning in what is, to the best of our knowledge, the largest think-aloud experiment ever run. Using a pipeline that codes participants' reasoning traces as search graphs, frontier LLMs display inter-rater reliability that is close to, but still less than, human-level. We then demonstrate the richness of think-aloud data by examining the generated search graphs and characterizing patterns in human reasoning traces. Thus, we argue this study offers a step towards realizing \citet{waterman1971protocol}'s vision of automated process-level analysis of human reasoning. 

\section{Method}

In our study, participants played the Game of 24, in which they are tasked with combining four starting numbers between 1 and 13 to make 24 using only basic arithmetic operations (addition, subtraction, multiplication and division). Participants in the think-aloud condition were asked to say all of their thoughts out loud as they solved the problem. Participants in the control condition were not prompted to verbalize. The preregistration for this experiment can be found \href{https://osf.io/3emh9/}{here}. Full code and data are available on \href{https://github.com/benpry/think-aloud-llms}{GitHub}.

\subsection{Participants}

We recruited 700 participants via Prolific, filtering for participants in the United States who were fluent in English, had at least a secondary education, and had an approval rate of at least 98\%. 640 participants were assigned to the think-aloud condition and 60 were assigned to the control. Participants took a median of 34 minutes to complete the experiment. They received a base payment of \$6.35 and earned a bonus of 10 cents for each problem they solved.

We excluded trials with transcripts that contained no information relevant to the task. If at least half of a participant's trials were excluded, we excluded the remaining trials as well. Of the 6400 trials we collected, 17 were excluded due to a technical error that corrupted the audio, 1240 were excluded due to a lack of relevant information, and 174 were excluded based on participant exclusions. We also excluded the 22 remaining trials from one of the problems, which was a repeat of a practice problem. This left us with 4,947 trials from 541 participants in the think-aloud condition.

\subsection{Stimuli}
We chose 200 problems of varying difficulty from \href{https://www.4nums.com/}{4nums.com}, which contains crowdsourced data on the difficulties of Game of 24 problems. To ensure that difficulties were similar across condition groups, we ordered the 200 problems by solve rate (taken from 4nums), and divided them into 10 deciles by difficulty. We then randomly sampled one problem from each decile, yielding 20 groups of 10 problems. 

\subsection{Procedure}

Participants in the think-aloud condition were randomly assigned to one of the 20 problem groups via counterbalancing, yielding 32 participants per group. We randomly assigned participants in the control condition to one of three problem groups, chosen from the 20 in the think-aloud condition. This gave us 20 participants per group.

Participants first read the instructions, which told them to try to say out loud everything that goes through their mind as they solve the task. Next, they completed two practice trials before continuing to the main experiment. The practice trials were the two easiest problems on 4nums. Each practice trial repeated until the participant answered correctly. 

Participants completed 10 main trials. In each trial, they entered equations on the screen using buttons that corresponded to the four starting numbers and the arithmetic operations. If a participant submitted an incorrect solution, they received a message informing them it was incorrect and the trial continued. Trials ended after the participant entered a correct solution, or after 3 minutes elapsed.

We used an automatic voice detector to ensure that participants were verbalizing their thoughts consistently. Every 20 seconds, it instructed the participant to verbalize more if the average vocal activity at the human voice range was below a threshold. After each trial, participants were asked to rate the extent to which they were saying their thoughts out loud on a 5-point Likert scale. They were asked to verbalize more of their thoughts if they responded 3 or lower. 

\section{Analysis pipeline}

After collecting data, we processed it using a pipeline that automatically transcribes recordings and turns them into graphs. This section describes the pipeline at a high level (see \href{https://github.com/benpry/think-aloud-llms}{repository} for additional details). 

\subsection{Transcribing recordings}

We transcribed the voice recordings using OpenAI Whisper large-v3, a state-of-the-art open-weights automated transcription model \citep{radford2023robust}. We used the hyperparameters that the authors report as best for long-form transcription, including beam search with 5 beams and temperature backoff. We also set a silence threshold of 20 seconds to skip long periods of silence.

274 trials had response times above 3 minutes due to errors. We truncated these trials' audio and marked them incorrect if they were above 181 seconds, allowing a second of lag.

\subsection{Filtering irrelevant transcripts}

Despite our best efforts to get participants to think aloud, some participants were either silent or did not say anything relevant to the task. We filter out these trials. We marked as irrelevant any trials where the transcript was a common string that Whisper outputs when tasked with transcribing silence, like ``Thank you.'' When transcripts did not exactly match one of these strings, we used the LLM Llama 3.3 70B \citep{grattafiori2024llama} with a prompt to determine whether the transcript contained any content relevant to the Game of 24. 

\begin{figure*}
    \centering
    \includegraphics[width=1\linewidth]{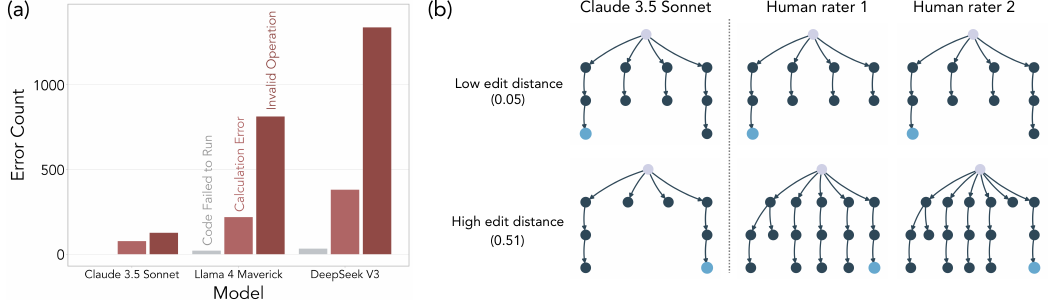}
    \caption{\textbf{Analysis of model-generated graphs}. (a) Counts of each type of automatically detected error for the three models that agree most with human coders. (b) Example graphs for trials with low and high mean human-model edit distance using Claude 3.5 Sonnet.}
    \label{fig:inter-rater-reliability}
\end{figure*}

\subsection{Coding transcripts as graphs}

In analyzing the data, we assume that participants mentally search for a solution by working through a series of possible game states connected by arithmetic operations.
Graphs are versatile data structures that are used extensively in search algorithms, so we use them to represent participants' search. Nodes in the graph represent game states---the sets of available numbers. Edges represent operations, which include combining any two numbers with an arithmetic operation or setting a subgoal. Subgoals are represented as backward edges from the goal state to the subgoal state. Figure~\ref{fig:overview}b shows example graphs. We also kept track of the order in which people explored operations.

Graphs were generated by prompting a language model to write Python code using a custom ``GraphBuilder'' class with methods designed to help build a search graph easily. Using graph representations requires the general theoretical commitment that human reasoning in this task can be formulated as a search process,  but we built only very general search operations into the class in order to avoid more specific theoretical commitments.

The language model had a system prompt containing instructions and the code for the custom class. We also included 10 in-context examples of transcripts with code translations written by the authors. Prompts also included the starting numbers for that trial, the response the participant submitted (if any), and the response time. We compared several LLMs for coding and used the results from the model that aligns best with human coding for subsequent analyses.

\paragraph{Automatic graph checking} After getting a translation from a language model, we ran the code and automatically checked the graph it generates for common errors, like computing the resulting state incorrectly or using numbers that do not occur in the current state.
When a graph had one of these problems, we called the language model again with a system prompt asking it to correct the code and in-context examples of code with errors and corrected code. We repeated this process up to five times, keeping the graph with the fewest problems. If a model failed to improve on a given iteration, we increased the temperature by 0.1 on the next iteration.

\section{Results}

Our analyses are organized around two goals: assessing the reliability and validity of the think-aloud method in our setting and examining characteristics of human reasoning traces.

\subsection{Reliability and validity}

\paragraph{Inter-rater reliability} We evaluated seven language models for their ability to code human transcripts: Claude 3.5 Sonnet, a high-performing proprietary model, as well as six open-weights models: DeepSeek V3-0324, Llama 4 Maverick and Scout, Qwen 3 235B-A22B, Llama 3.3 70b, and Llama 3.1 8b \citep{anthropic2024claude,liu2024deepseek,qwen2025qwen,grattafiori2024llama}. Performance was measured computing the average agreement with human coders. Two of the authors manually annotated a set of 50 randomly chosen trials, using the same graph checker and prompt as seen by LLMs, as well as a graph visualization function the LLMs did not have access to. We then computed the normalized graph edit distance between the graphs created by human and language model coders using the networkx Python package \citep{hagberg2008exploring,abu2015exact}.\footnote{Graph edit distances are computed where nodes can only be matched to each other if they correspond to the same state. We normalize the graph edit distances by dividing them by the sum of the maximum vertex and edge counts. That is, $\max(|V_1|, |V_2|) + \max(|E_1|, |E_2|)$ Code that does not run is given an edit distance of 1.} 

As is shown in Figure~\ref{fig:overview}c, agreement between human coders and language models is somewhat lower than inter-rater reliability between the human coders, with all language models exhibiting significantly less agreement with the humans than the humans do with each other ($p < 0.002$ in all cases, computed via permutation test). However, there is substantial overlap between the distributions of human-model and human-human edit distances, particularly for Claude 3.5 Sonnet, DeepSeek V3, and Llama 4 Maverick. Moreover, the model-produced graphs generally look sensible and qualitatively similar to human graphs, and capture important features like the path from the root node to the goal even at high disagreement (see Figure~\ref{fig:inter-rater-reliability}b). The human coders took on average 8 hours and 15 minutes to annotate 50 trials, which amounts to 816 hours to code all 4,947 trials at the same pace. In contrast, coding all trials with Claude took less than 2 hours and cost \$74. While graphs produced by the models should be interpreted with some caution, they are close enough to human-generated graphs to be informative about human reasoning and enable analysis of reasoning traces at a scale that is not feasible using manual coding. 

\paragraph{Error analysis} We coded the full dataset with the three models that performed best in terms of graph edit distance: Claude 3.5 Sonnet, DeepSeek V3, and Llama 4 Maverick. After five iterations of automatically checking the graphs for errors and correcting them, some of the graphs still had errors. Figure~\ref{fig:inter-rater-reliability}a shows the total counts of each of the three types of errors that occurred: non-runnable code, incorrect resulting states, and operations that use numbers that do not exist in the states they start in. Claude produced the fewest errors, and never produced non-runnable code, so we use its annotations for our remaining analyses.

\paragraph{Effect of prompting to think aloud} We compared participant performance between the think-aloud condition and the control condition to test whether thinking aloud affects performance. Participants in the control condition found the correct solution in 53\% of trials, compared with 49\% of trials on the same problems in the think-aloud condition. This difference came close to significance according to a permutation test ($p = 0.067$). A follow-up analysis found that participants in the think-aloud condition took significantly longer to respond on average (125.1 vs. 117.7 seconds, $p = 0.008$). The effect of thinking aloud on response times was similar across problems, as evidenced by a strong correlation between average problem-level response times across conditions ($r=0.95,\ p < 0.0001$).

\begin{figure*}[]
    \centering
    \includegraphics[width=1\linewidth]{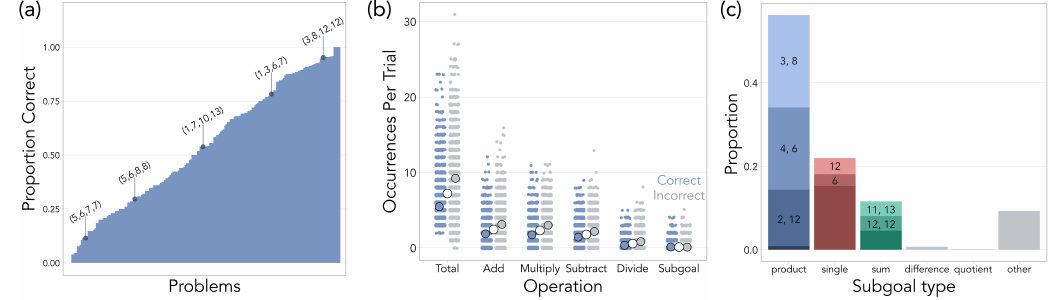}
    \vspace{-0.5cm}
    \caption{\textbf{Summary of participant data}. (a) Proportion of correct answers for every problem in the dataset. (b) Mean number of uses of each operation in trials where the participant was correct and incorrect. (c) Proportion of subgoals that are of each type, defined by how one can reach the goal from them.}
    \label{fig:graph-features}
\end{figure*}

We also tested whether thinking aloud influences performance on some problems more than others—which would suggest a change to the underlying reasoning processes. We computed split-half correlations for mean accuracy by problem in the think-aloud condition by randomly dividing the data into pairs of groups with 30 participants each (10 per stimulus group). We then compared the correlations between these groups to correlations between the think-aloud and control conditions, where participants were downsampled in the same way. We found no significant difference ($r=0.83$ within think-aloud, $r=0.79$ between conditions, $p = 0.74$).

Taken together, these results indicate that while asking participants to think aloud may slow them down, there is no evidence that it changes the pattern of which problems are easy and hard. This is consistent with findings from a prior meta-analysis \citep{fox2011procedures}. The large scale of our data enables us to robustly measure differences in overall performance and correlations between problem-level difficulties, which would not have been possible in a smaller study.

\subsection{Characteristics of human reasoning traces}

We next analyze aggregate patterns in how people search, including how often people use each type of operation, the extent to which they cluster around certain sequences of operations, and what leads to successful and unsuccessful search.

\paragraph{Use of operations} A natural first question to ask is how often people consider each type of operation.
Figure~\ref{fig:graph-features}b shows the mean number of times each type of operation occurs in participants' search graphs, for trials where the participant got the answer correct and incorrect. Addition is the most common operation, followed closely by multiplication and subtraction, with division a distant fourth. This may indicate that in our task division is less cognitively natural to participants than other arithmetic operations. Participants explored more operations in incorrect trials, generating significantly more edges of all types except for subgoals. In successful trials, participants were able to efficiently reach the goal while considering relatively few operations. When participants did not get the right answer, they continued to try more operations.

Setting subgoals was rare: only 445 trials (9.0\%) contain one or more subgoals and 58\% of participants set no subgoals at all. Figure~\ref{fig:graph-features}c shows the frequencies of the different types of subgoals, along with some of the most common individual states. Product subgoals, consisting of two factors of 24, were the most common by far. Many of the subgoals were single-number states. While it is not possible to reach 24 from these states, they tend to occur when the participant tries to make a number without mentioning the other numbers they would need to get to the goal, like ``Can I make 12?'' 12\% of the subgoals were sum subgoals and almost none were differences or quotients. While participants explored addition and multiplication about equally, they set far more product subgoals than sum subgoals. Setting a product subgoal may be easier because there are only four natural number-valued states that could lead to the goal by multiplying, while there are 24 that could lead to the goal by adding.

\begin{figure*}[]
    \centering
    \includegraphics[width=\linewidth]{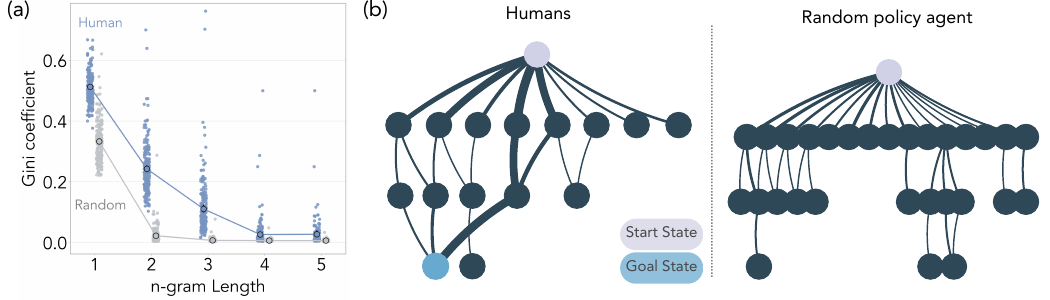}
    \caption{(a) Gini indices for sequences of operations of lengths 1 through 5 in humans (blue) and random search agents (gray). Each dot represents one Game of 24 problem. (b) Aggregate visualizations of human and random reasoning traces for one problem. Edge widths represent the number of times an operation was explored.  Operations only explored once are filtered.}
    \label{fig:ngram-ginis}
\end{figure*}

\paragraph{Consistency and variation in search patterns} Next, we analyze the extent to which participants use similar operations, and sequences of operations, to each other using Gini indices. This analysis follows \citet{mccarthy2023consistency}. The Gini index is a measure of dispersion, where higher numbers indicate more clustering: a value of 1 would mean all subsequences were identical, while a value of 0 would mean all subsequences were unique.\footnote{Due to the size of the search space, Gini indices were computed over only subsequences that occurred at least once in the data.}

Figure~\ref{fig:ngram-ginis}a shows the Gini coefficient for subsequence lengths 1-5 in humans and an agent that explores random legal actions matched to the humans in number of operations explored. Human search clusters much more around particular subsequences than the random agent, particularly for subsequences of lengths 2 and 3. This indicates that humans exhibit consistent multi-step reasoning strategies that extend beyond single operations.

\paragraph{Predictors of success and failure in reasoning} We can learn about people's reasoning strategies from when and how they succeed and fail at the task. Participants struggled with problems requiring division. The 29 problems in the experiment that could only be solved using division had a success rate of 20\%, much lower than the 59\% success rate for problems that can be solved without division ($p < 0.0001$). 47\% of participants who got those problems wrong never tried di-
vision, suggesting that failures of consideration
are a main driver of participants’ reasoning failures.

Despite subgoals being rare overall, they appear to be helpful. Participants were significantly more likely to find the correct answer in trials where they set one subgoal than trials where they set none (65\% vs. 53\%, $p < 0.0001$). However, they were no more likely to find the correct answer when they set more than one subgoal than when they set none (47\% vs. 53\%, $p = 0.35$). This result suggests that working backward can help, even though people rarely do it. Setting multiple subgoals might indicate that a participant struggled with a problem and failed to achieve their first subgoal.\footnote{These patterns were either weaker or non-significant using DeepSeek V3 and Llama 4 Maverick, yet note these models show substantially higher error rates compared to Claude 3.5 Sonnet.}

\section{Discussion}

Reasoning is a hallmark of the human mind, but it is difficult to study from performance alone. The think-aloud method is a classic approach to eliciting information about the process of reasoning, but it has historically been limited by the amount of manual effort involved in coding.In this paper, we developed methods to automate think-aloud analysis in order to study human reasoning with both granularity and scale. We ran a large-scale experiment with 700 participants playing the Game of 24 and analyzed the data using an automated pipeline. We validated our methods by comparing them to human annotations. While the model reliability was lower than humans, it was high enough to be a favorable tradeoff given the substantial time and cost savings. From the graph representations, we learned that participants cluster around similar parts of the search space and discovered that people use addition and multiplication most often. Failing to consider division seemed to be a substantial impediment to participants' success. Our results illustrate the utility of think-aloud studies in gathering process-level data about reasoning.

Our automated pipeline overcomes the issue of scale, but our study still inherits some limitations of traditional think-aloud experiments. For instance, the verbal reports may not include everything a participant considers \citep{wilson1994proper,ericsson1993protocol}, and our analysis pipeline may not have captured all of what they said. Secondly, our graph representations cannot express all of the participants' thoughts. For instance, ``There has to be some multiplication here,'' does not fit into any of our graph operations. We used simple representations to study the operations people use, but different questions might require different representations. One potential solution to this is to employ more open-ended code generation from LLMs, as was done in a recent study in which an LLM generated code from participants' retrospective descriptions of their strategy in a sorting task \citep{xie2024strategic}.

Automating think-aloud analysis enables us to overcome major limitations of traditional think-aloud studies. After an initial investment in the coding scheme and manually annotated examples,  we were able to automatically code thousands of trials. This would simply not be feasible with manual coding. This automated method is also more reproducible than manual coding, as sharing prompts and in-context examples enables other researchers to obtain the same results. When using a closed-weights model, reproducibility is contingent on the model remaining available through an API. However, as open-weights models continue to improve, we expect reliable fully-open coding pipelines to be possible soon.

We hope that this work serves as a proof of concept that inspires the future development of automated methods for analyzing think-aloud data. We see the rich, granular data enabled by the think-aloud method at scale as a promising avenue for gaining new insights into how humans reason.

\bibliographystyle{apacite}

\setlength{\bibleftmargin}{.125in}
\setlength{\bibindent}{-\bibleftmargin}

\section{Acknowledgments}

The authors thank the Stanford Computation and Cognition Lab for feedback on early versions of this work. Particular thanks to Linas Nasvytis, Allison Chen, Samah Abdelrahim, and Veronica Boyce for their helpful comments on a draft of this paper. KG was supported by an HAI-SAP Grant and an NSF Expeditions grant.

\bibliography{citations}

\end{document}